# Advanced deep-reinforcement-learning methods for flow control: group-invariant and positional-encoding networks improve learning speed and quality

Joongoo Jeon[a,b*], Jean Rabault[c], Joel Vasanth[d], Francisco Alcántara-Ávila[d], Shilaj Baral[a], Ricardo Vinuesa[d]

jgjeon41@jbnu.ac.kr, jean.rblt@proton.me, jvasanth@kth.se, fraa@kth.se, shilajbaral@jbnu.ac.kr, rvinuesa@mech.kth.se

[a]Graduate School of Integrated Energy-AI, Jeonbuk National University

[b]Department of Quantum System Engineering, Jeonbuk National University

567 Baekje-daero, Deokjin-gu, Jeonju 54896, Republic of Korea

[c]Independent Researcher, Oslo, Norway

[d]FLOW, Engineering Mechanics, KTH Royal Institute of Technology

SE-100 44 Stockholm, Sweden

*Corresponding author: Joongoo Jeon (jgjeon41@jbnu.ac.kr)

**Abstract**

Flow control is key to maximize energy efficiency in a wide range of applications. However, traditional flow-control methods face significant challenges in addressing non-linear systems and high-dimensional data, limiting their application in realistic energy systems. This study advances deep-reinforcement-learning (DRL) methods for flow control, particularly focusing on integrating group-invariant networks and positional encoding into DRL architectures. Our methods leverage multi-agent reinforcement learning (MARL) to exploit policy invariance in space, in combination with group-invariant networks to ensure local symmetry invariance. Additionally, a positional encoding inspired by the transformer architecture is incorporated to provide location information to the agents, mitigating action constraints from strict invariance. The proposed methods are verified using a case study of Rayleigh-Bénard convection, where the goal is to minimize the Nusselt number $Nu$. The group-invariant neural networks (GI-NNs) show faster convergence compared to the base MARL, achieving better average policy performance. The GI-NNs not only cut DRL training time in half but also notably enhance learning reproducibility. Positional encoding further enhances these results, effectively reducing the minimum $Nu$ and stabilizing convergence. Interestingly, group invariant networks specialize in improving learning speed and positional encoding specializes in improving learning quality. These results demonstrate that choosing a suitable feature-representation method according to the purpose as well as the characteristics of each control problem is essential. We believe that the results of this study will not only inspire novel DRL methods with invariant and unique representations, but also provide useful insights for industrial applications.

## 1. Introduction

A precise regulation of the fluid-dynamics characteristics and the thermal gradients through advanced flow- and heat-control systems has a decisive impact on improving energy efficiency and safety. This importance can be easily gauged from examples across many energy industries. For instance, countries around the world are racing to develop cutting-edge design concepts for small modular reactors [1]. Many researchers have investigated helium bubbling control systems for enhancing natural circulation in molten-salt fast reactors [2]. For wind turbines, active-flow-control technologies play a significant role in optimizing the aerodynamic performance of turbine blades [3]. Techniques such as flow-field management and active-control methods, which adjust the flow characteristics over the blades, help to maximize energy capture while minimizing wear and tear on the equipment. This leads to higher efficiency and extended lifespan of wind turbines, which is crucial for cost-effective renewable-energy production [4]. However, the application of these flow-control systems is limited because traditional control algorithms exhibit problems when dealing with high-dimensional non-linear operating raw data [5]. The proposed methods in this study aim to overcome these limitations by integrating advanced DRL frameworks with domain-specific knowledge, making it feasible to implement efficient and robust flow control in various energy systems.

DRL, which leverages the strengths of deep learning and RL to solve complex problems, emerges as a promising technique in flow control [6]. This technology recently been utilized across various physics and engineering fields to address decision-making problems that were previously unsolvable due to their non-linear nature and high dimensional complexity [5, 7]. RL is a branch of artificial intelligence where an agent interacts with an environment to learn optimal actions through trial and error [8]. As shown in **Fig. 1**, the agent is responsible for making decisions within the environment, taking actions that influence the state of the environment. The environment represents the external system on which the agent operates, and its state corresponds to the current values of the environment's variables. Actions are the decisions made by the agent to transition from one state to another. The agent receives feedback in the form of rewards or penalties based on the actions taken, providing a signal to guide its learning. The policy is the strategy or set of rules the agent employs to determine its actions in different states. Unlike the way supervised and unsupervised learning techniques are applied in the engineering field [9, 10], the goal of RL is for the agent to learn a policy that maximizes the cumulative or instantaneous reward over time [8].

The power of DRL lies in its ability to learn a policy in complex and dynamic environments, facilitated by neural networks that effectively approximate complex non-linear, high-dimensional value functions or policy distributions [6]. This approach has led to significant breakthroughs across various fields, such as robotics [11], finance [12], and flow control [5, 6]. In particular, since the proximal policy optimization (PPO) algorithm implemented by Rabault et al successfully led to the first control of a flow [6], DRL algorithms are extensively researched in active flow control (AFC) problems due to their promising engineering potential. **Table 1** summarizes a few of the recent studies on DRL applications to flow control. Zhang et al. [13] Wang et al. [14]

Table 1. Summary of related studies applying RL for flow control

| Year | Author | Dataset type | Algorithm | Key ideas |
| --- | --- | --- | --- | --- |
| **2019** | Rabault et al. [15] | Kármán vortex | PPO | First application of neural networks |
| **2019** | Rabault et al. [16] | Kármán vortex | PPO | Multi-environment approach |
| **2019** | Belus et al. [17] | Falling Liquid film | PPO | Translational invariance |
| **2020** | Tang et al. [18] | Synthetic jets | PPO | DRL application for drag reduction |
| **2021** | Zhang et al. [13] | DMC | SAC | Invariant representation for acceleration |
| **2021** | Zeng et al. [19] | KS | DDPG | Invariance by Fourier transformation |
| **2021** | Radaideh et al. [20] | Nuclear assembly | PPO/DQN | Physics-informed RL |
| **2022** | Wang et al. [14] | PyBullet | DQN/SAC | Equivariant RL |
| **2023** | Li et al. [21] | Steam generator | PPO | Nuclear energy applications |
| **2023** | Vignon et al. [22] | RBC | PPO | Translational invariance by multi-agent |
| **2023** | Guastoni et al. [23] | Channel flow | DDPG | Comparison with opposition control |
| **2023** | Suàrez et al. [24] | 3D Cylinder | PPO | Three-dimensional control |
| **2024** | Peitz et al. [25] | KS equation, etc. | DDPG | Equivariance by convolutional RL |
| **2024** | Font et al. [26] | TSB | PPO | Three-dimensional control |

Despite the advancements in DRL technology, there are practical difficulties in applying it to flow control in energy systems. During the training procedure, DRL agents need to explore a wide range of actions and states to find the optimal policy [27]. This exploration can be computationally expensive, especially in environments with a large state-action space [28]. Furthermore, unlike supervised learning where pre-constructed data can be used, DRL generally involves online learning where the agent updates its parameters based on each new interaction with the environment [8]. Each new interaction requires a new simulation of the environment, which must be built in advance according to the trajectory of each episode. In complex environments, DRL often converges to a sub-optimal policy and cannot recover [27]. Flow-control simulations of energy systems represent complex environments due to their high-dimensional flow fields [6]. Consequently, it is essential to design an RL method that allows the agent to gain as much knowledge as possible about the environment in one episode, reducing costs and enhancing optimal policy performance. This necessity has driven recent research on accelerating RL by utilizing equivariance and invariance, concepts that help improve learning efficiency [19]. The difference between equivariance and invariance is explained in detail in *Section 4*.

The objective of this study is to develop advanced DRL methods by integrating domain-specific knowledge about the target control problem into the design of the DRL setup. In particular, we can expect that an invariant operator considering the properties of the flow and action in the control system can improve DRL performance in the reduced subspace. Indeed, we can reduce the complexity of the problem by taking invariants implied by the equations into account, the agent can then explore policies more effectively. In other words, the combinatorial cost explosion caused by the curse of dimensionality on the action space dimensionality can be mitigated by taking advantage of inherent structure in the underlying system. Extending on the MARL setup to exploit translational invariance, we list on the following major contributions of the present work:

(1) we develop two types of group invariant networks that also exploit symmetrical invariance
(2) we develop a unique representation method to alleviate conflicts between agent actions in invariant environments.

The group-invariant networks base their foundation on the group theory and group-equivariant network [29, 30], while the positional encoding is adopted from the transformer model, which has been successful in sequential data processing [31]. These advancements aim to improve the effectiveness of DRL in controlling complex non-linear, high dimensional fluid-flow systems by leveraging both inherent invariants and symmetries, combined with effective neural network representation techniques.

As shown in **Table 1**, a few DRL research studies applying feature representations have been conducted in the past. For example, Vignon et al. [22] and Belus et al. [17] developed a translational invariant DRL framework using a multi-agent method to increase learning efficiency. This method has further been adapted to several other flow control situations, with great success [23, 24]. However, the MARL invariant methodology per se is limited to representing translation invariance. Extending on this, Zeng et al. applied symmetry-reduced feature representation using Fourier subspace [19]. Their methodology is very innovative, but it is limited to cases where the solution can be naturally expressed in terms of Fourier modes. In the field of artificial intelligence, research on invariant/equivariant RL is being actively conducted such as the research of Zhang et al. [13] and Wang et al [14]. To the best of the authors' knowledge, our present work is the first study to investigate the efficacy of group-invariant networks and positional encoding on DRL by developing a general-purpose network architecture that can be applied to fluid-flow systems. To evaluate the performance of this study, the reduction of the Nusselt number $Nu$ in two-dimensional Rayleigh-Bénard convection (RBC) is selected as a case study.

The rest of this paper is organized as follows. *Section 2* addresses the baseline control framework with DRL. *Section 3* describes our strategy to enhance DRL performance in more detail. *Sections 4 and 5* introduce the concept and performance of the developed group-invariant neural networks and positional encoding approach, respectively. *Section 6* summarizes and concludes this paper.

## 2. RBC control framework with reinforcement learning

### 2.1 Fundamentals of RL

Reinforcement-learning methods are primarily classified into value-based RL and policy-based RL [8]. The value of a state, or 'value function' is the expectation of the cumulative rewards (also called the expected return) of all future states from that state. Value-based RL involves estimating the value function and using these estimates to inform decision-making. The learned value function is used to derive an optimal policy [8]. Conversely, policy-based RL directly parameterizes and optimizes the policy without explicitly learning a value function [8]. The policy determines the action or a distribution over the actions given a state. Policy-based RL is particularly effective in continuous or high-dimensional action spaces, making it suitable for flow control [32]. However, a significant disadvantage of the policy-based approach is that it often requires more samples to achieve good performance [33]. Studying algorithms to improve learning efficiency is essential in RL, where obtaining data is a time-consuming online process. Unlike supervised learning, RL usually gathers data through online, close-loop trial-and-error interaction with the environment.

Policy-gradient methods, one of the representative methods in policy-based RL, are further categorized into stochastic policy-gradient RL and deterministic policy-gradient RL [8]. Stochastic policy-gradient methods optimize a policy that outputs a probability distribution over actions, with actions sampled from this distribution during training and inference. Deterministic policy-gradient methods optimize a policy that deterministically maps states to actions, with actions directly obtained by the policy without sampling from a distribution. In this study, PPO algorithm, classified as a stochastic policy-gradient RL method, is applied to RBC control. The PPO method is a significant improvement over the former TRPO method [34, 35], in terms of the nature of the constraint imposed on the magnitude of successive policy updates: in PPO, the policy update constraint used is a clipping function, and is embedded directly into the loss function, making it an unconstrained optimization problem. In TRPO, a KL-divergence constraint on policy updates appears independently of the loss function, as in a constrained optimization problem. Further, the constraint in TRPO requires second order derivatives (Hessian vector-products) to solve which are computationally expensive to obtain. PPO on the other hand requires first order derivatives only. PPO uses a clipped surrogate objective defined as follows:

$$L^{CLIP}(\theta) = \hat{E}_t\left[min(r_t(\theta)\hat{A}_t, clip(r_t(\theta), 1-\epsilon, 1+\epsilon)\hat{A}_t\right], \tag{1}$$

where $r_t(\theta) = \frac{\pi_\theta(s_t)}{\pi_{\theta_{old}}(s_t)}$ is the probability ratio, $\hat{A}_t$ is the advantage estimate, $s_t$ is the state at time $t$, and $\epsilon$ is a hyperparameter determining the clipping range [35]. Trust-region policy optimization (TRPO) employs a more complex optimization technique involving conjugate-gradient methods to solve the constrained optimization problem, which includes computing the Hessian-vector product and is more computationally intensive [34]. As a stochastic policy-gradient method, PPO naturally balances exploration and exploitation, which is crucial in flow-control tasks to discover efficient strategies without prematurely converging to suboptimal policies [35]. Additionally, PPO is relatively easy to implement and tune compared to other advanced policy-gradient methods, facilitating experimentation and deployment in flow-control applications [15]. A more detailed description of PPO is provided in Algorithm 1 and in Refs. [35].

**2.2 RBC baseline control framework**

To study the effectiveness of the RL methods developed in this study, we investigate the 2D RBC control problem. This control problem, which involves both momentum and heat transfer, allows verification of the RL applicability in energy and fluid-flow systems. Therefore, this allows us to test our ideas and implementations on a simple test case that is computationally affordable while containing the properties of invariance, equivariance, and symmetry that we want to study. We use the Tensorforce framework, a TensorFlow library for applied reinforcement learning, for PPO implementation [36, 37]. **Fig. 1** illustrates the base MARL control methodology for RBC, integrated with a computational-fluid-dynamics (CFD) solver. This baseline setup is largely similar to [22]. The objective of this problem is to minimize the Nusselt number *Nu* by regulating the temperature of the lower wall, which is divided into 10 segments. The figure depicts the MARL framework by displaying 10 pseudo-environments. The pseudo-environment is an environment in which each agent interacts independently, and its number is equal to the number of segments (or agents) $N_s$ in this multi-agent framework. Control invariance across the domain in the MARL setup is achieved by sharing the same policy (same parameterized neural networks) among all agents in pseudo-environments, enabling to share experience among MARL environments, which allows efficient learning and control. This approach mitigates the curse of dimensionality by breaking down the global control problem into smaller localized tasks [22]. In other words, with one global simulation, we can get 10 times more parameter update information compared to a single-agent approach. The subscripts for state $s$, action $a$, and reward $r$ are indexed by the environment (or segment) index. The state $s_i$ represents the observation values from probes above control segment $i$ and consists of temperature, $x$-axis velocity, and $y$-axis velocity. The value $r_i$ is the total reward from pseudo-environment $i$, and $a_i$ is the control action, $i$ goes from 1 to $N_s$.

The MARL methodology is now described. We begin from the set of all actions output by the agents. Initially, the actions $a_i$ are combined into a single vector and communicated to a single pseudo-environment, in this case, the first pseudo-environment. Next, the first pseudo-environment runs the numerical simulation, while the other pseudo-environments wait for its completion. After the simulation, the full observation vector is retrieved and communicated to all the pseudo-environments. Thus, each pseudo-environment now has a copy of the observation vector. A 'recenter()' function rearranges the order of the observation vector for each pseudo-environment, such that $s_i$ appears in the center of the observation vector for that environment. The modified observations (recentered state $s_i'$) and rewards are then passed to the MARL agent to output the next actions, repeating the cycle [22]. The advanced DRL methods developed in this study are based on this baseline control framework.

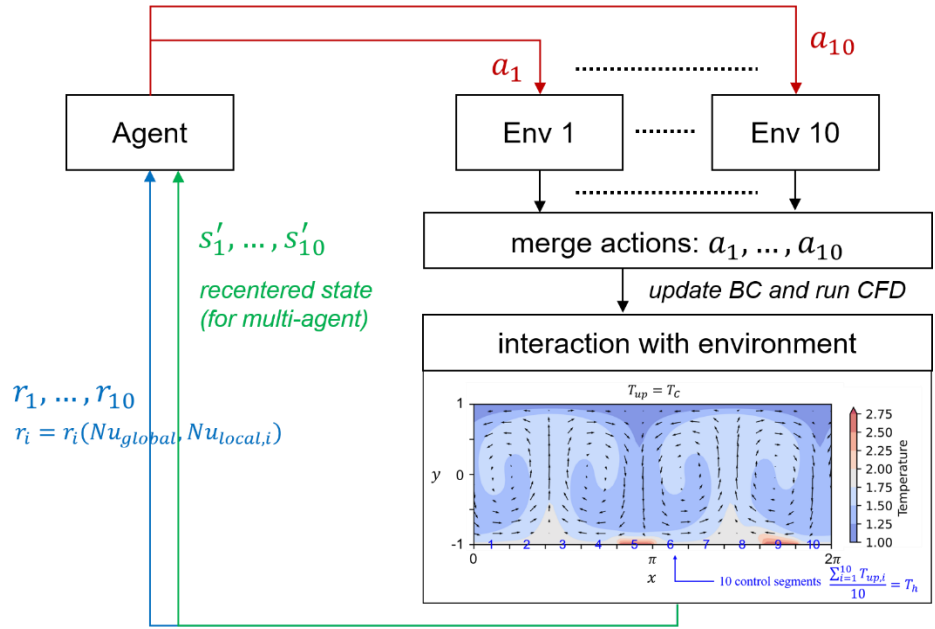

**Fig. 1.** MARL framework for RBC control using a CFD solver. The environment shows natural convection occurring between the upper wall with constant temperature and the bottom wall with a temperature distribution that varies based on control actions. More detailed descriptions of the computational domain normalized by $H$ and the multi-agent approach can be found in [22].

**Algorithm 1** outlines the overall DRL framework. The reward function is composed of two values of $Nu$, which is defined as $Nu = \frac{q(t)}{\kappa(T_H - T_C)/H}$ as shown in **Eq. (2)**. As noted in Ref. [22], the global $Nu$ incentivizes each pseudo-environment to improve the global flow state, while the local *Nu* provides more reward granularity to the agent during training by generating a local indication of the quality of control.

$$r_i = m(n - (1-\beta)Nu_{global} - \beta Nu_{local,i}), \tag{2}$$

Note that $m, n$ are DRL hyperparameters and $\beta$ is the weighting factor, 0.0015 [22]. In this framework, actions interact with the environment through temperature changes in each segment, denoted as $a_i = T'_i$. The raw action is computed within the interval [-1, 1]. We impose a constraint on the mean temperature over the bottom wall of the domain. This is because deviations from a constant mean temperature would mean a change in the overall heat flux across the domain, modifying the regime of the instability. This is done to avoid the trivial control strategy of lowering the overall bottom wall temperature to reduce convection. We thus convert raw actions to temperature actuations as follows. First, we adjust each action , based on the mean temperature of the raw action, as shown in **Eq. (3)**. Additionally, the maximum absolute value of the final action is constrained to $T'_i < |0.75|$ [22].

$$a_i^f = a_i - \frac{\sum_{j=1}^{N_s} a_j}{N_s}, \tag{3}$$

where the final action $\boldsymbol{a}_i^f$ changes the temperature of the $i$-th segment in the next state by that amount.

---

**Algorithm 1** PPO-Clip in MARL framework

Input: initial policy parameters $\boldsymbol{\theta}_0$, initial value function parameters $\boldsymbol{\phi}_0$
**for** $k = 0, M$ **do**
  **for** $t = 0, T$ **do**
    Execute merged action $\boldsymbol{a}_t$
      **for** $i = 0, N$ **do**
        Recentering $s_{i,t} \leftarrow s'_{i,t}$
        Compute rewards-to-go $r_{i,t}$ and current value function $V_\phi(s_{i,t})$
        Compute advantage estimates $\hat{A}_{i,t}$ with recentered state $s_{i,t}$
        Compute action $a_{i,t}$ following $\pi_\theta$ with recentered state $s_{i,t}$
  **end for**
  **end for**
Optimize surrogate $L$ wrt $\boldsymbol{\theta}, \boldsymbol{\phi}$

$$\boldsymbol{\theta}_{k+1} = \underset{\theta}{\operatorname{argmax}} \frac{1}{|D_k|T} \sum_{\tau \in D_k} \sum_{t=0}^{T} \sum_{i=0}^{N} \min\left(\frac{\pi_{\theta(a_{i,t}|s_{i,t})}}{\pi_{\theta_k(a_{i,t}|s_{i,t})}} A^{\pi_{\theta_k}}(s_{i,t}, a_{i,t}), g\left(\epsilon, A^{\pi_{\theta_k}}(s_{i,t}, a_{i,t})\right)\right)$$

$$\boldsymbol{\phi}_{k+1} = \underset{\phi}{\operatorname{argmin}} \frac{1}{|D_k|T} \sum_{\tau \in D_k} \sum_{t=0}^{T} \sum_{i=0}^{N} (V_\phi(s_{i,t}) - r_{i,t})^2$$

  via gradient descent algorithm
**end for**

## 2.3 Numerical method for CFD

The numerical method for the RBC simulation in this study employs a Shenfun platform based on the method developed by Kim, Moin, and Moser for direct numerical simulations of turbulent flows [38]. Shenfun is a high performance computing platform for solving partial differential equations (PDEs) by the spectral Galerkin method [39, 40]. This CFD method eliminates the pressure term and reduces the two-dimensional Navier-Stokes equations to the continuity equation and a fourth-order equation for the wall-normal velocity component. Finally, the RBC can be solved using the continuity equation (**Eq. (4)**), the momentum equation (**Eq. (5)**), and the energy equation (**Eq. (6)**) [38].

$$\nabla \cdot \mathbf{u} = 0, \tag{4}$$

$$\frac{\partial \nabla^2 v}{\partial t} = \frac{\partial^2 H_x}{\partial x \partial y} - \frac{\partial^2 H_y}{\partial x^2} + \sqrt{\frac{Pr}{Ra}} \nabla^4 v + \frac{\partial^2 T}{\partial x^2}, \tag{5}$$

$$\frac{\partial T}{\partial t} + \mathbf{u} \cdot \nabla T = \frac{1}{\sqrt{RaPr}} \nabla^2 T, \tag{6}$$

where $u(x, y, t)$ and $v(x, y, t)$ is the velocity of $x$-axis and y-axis, $T(x, y, t)$ is the temperature. $\mathbf{u}$ is the velocity vector. The Prandtl number is defined as $Pr = \frac{\nu}{\kappa}$, the Rayleigh number is defined as $Ra = \frac{\alpha(T_H - T_C)gH^3}{\kappa \nu}$, and $H = (\mathbf{u} \cdot \nabla)\mathbf{u}$ represents the convection vector. Boundary conditions include no-slip conditions at the walls and periodic conditions in the horizontal direction. The spectral Galerkin method employs tensor-product basis functions from Chebyshev polynomials for the wall-normal direction and Fourier exponentials for the periodic direction, ensuring exact enforcement of boundary conditions. The numerical implementation leverages the Shenfun open-source spectral Galerkin framework, allowing automatic discretization through high-level scripting. To avoid aliasing, the convection terms are computed in physical space after expanding the number of collocation points [40].

The accuracy of the Navier-Stokes solver is verified by reproducing the growth of the most unstable eigenmode of the Orr-Sommerfeld equations over long time integrations. The framework and solvers are publicly available, with detailed descriptions provided in the documentation, ensuring reproducibility and facilitating further evaluations of various domain sizes and non-dimensional parameters [39, 40]. The solver setup and CFD-DRL coupling used in the present work are direct extensions based on the open source code released by Vignon et. al (https://github.com/KTH-FlowAI/DeepReinforcementLearning_RayleighBenard2D_Control) [22].

## 3. Methods

**Fig. 2** illustrates DRL concepts using a simplified depiction of the flow field. Vignon et al. verified that the translation-invariant properties inherent to the MARL framework significantly increases DRL capability **(Fig. 2(b))** [22]. Compared to the single-agent method (**Fig. 2(a)**), the convergence speed, and the quality of the converged policy was improved [22]. However, the MARL method does not guarantee a rotation-invariant or symmetry-invariant policy. Based on our domain knowledge about RBC, we can intuitively expect that when a state is symmetric, the required action to reach a controlled state will also be symmetric. Indeed, the underlying equations are symmetric too, and hence the solution of the problem will be symmetry, too. **Fig. 3(a)** shows a symmetry-invariant example of state-dependent action in a simple flow field through a 4 segment MARL framework. In three-dimensional environments, the invariant property is similarly required for rotation transformations. If we make it possible for the MARL method to learn a policy that is invariant to rotation (or symmetry) as well as translation, we could expect the performance of DRL to increase further, since we embed more of the problem structure into the DRL setup.

Interestingly, efforts to enhance machine-learning performance for rotational transformations, as well as translational transformations, have been ongoing in computer science [29, 30]. Since convolutional correlation is translation-equivariant but not rotation-equivariant, new CNN-based architectures for rotation-equivariant capabilities have been developed. Among these, group convolutional neural networks (G-CNNs) are one of the most successful architectures [29, 30]. However, in flow control applications such as RBC, a rotation-invariant architecture is required rather than a rotation-equivariant one as shown in **Fig. 3**. Rotation-invariant relying on pooling layers makes it difficult to predict continuous actions, unlike classification. In this study, as shown in **Fig. 2(c, d)**, we develop a novel architecture for rotation invariance (or symmetry invariance) based on the G-CNN architecture. More detailed descriptions of group-invariant neural networks are provided in *Section 4*.

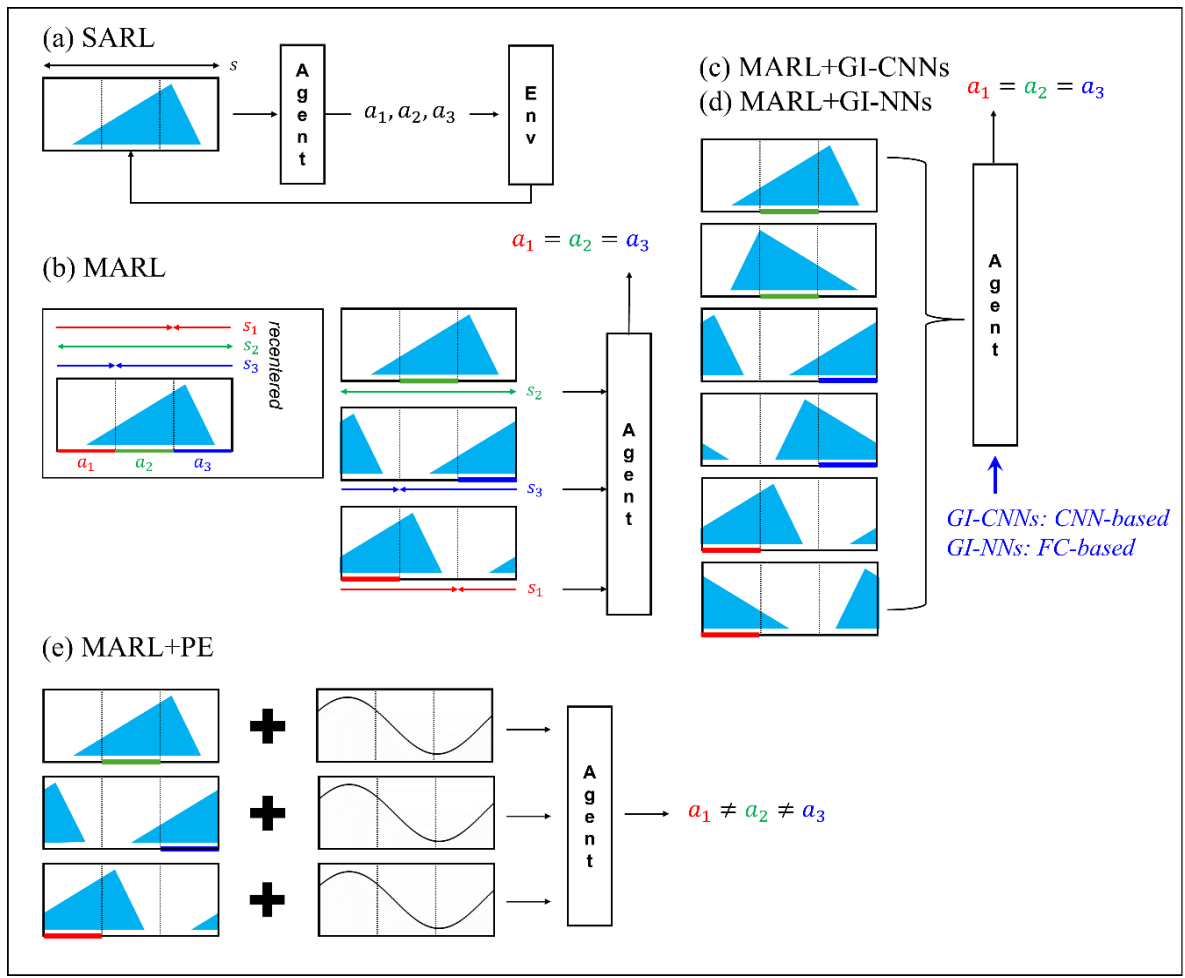

**Fig. 2.** Illustrations of base and new DRL models based on simplified flow field (left/right walls: periodic boundary conditions). (a) SARL (b) MARL (base DRL-RBC framework) (c) GI-CNNs model (d) GI-NNs model (e) PE-NNs model. Each model performed different feature representations by imposing our domain knowledge about the RBC control.

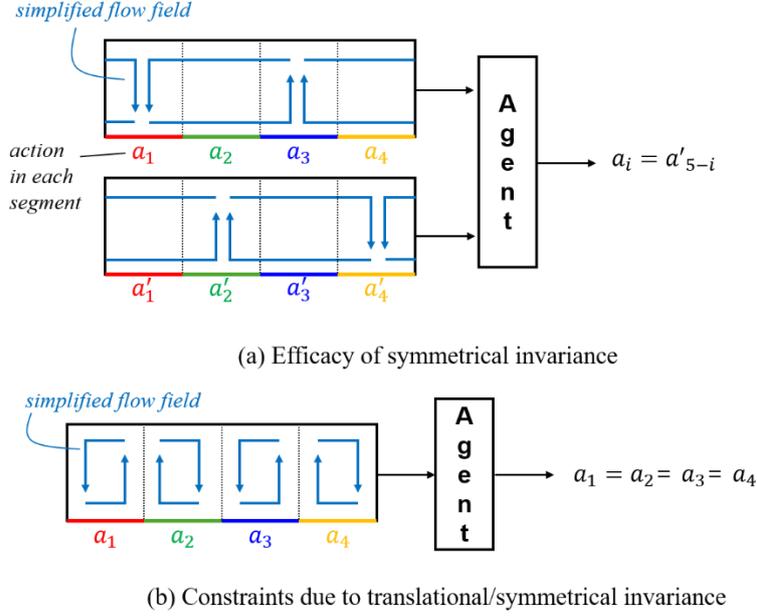

(a) Efficacy of symmetrical invariance

(b) Constraints due to translational/symmetrical invariance

**Fig. 3.** Illustration showing the obtained insights from phenomenological analysis of RBC flow. On the left is the RBC domain, the blue arrows are a simplification of the manner of rotation of the velocity field during RBC instability. Our first strategy is to impose symmetrical invariance on the DRL framework as in (a). However, as in (b), strict translation/symmetry invariance can constrain the agent's effective actions. Our second strategy is to provide the agent with location information to relax action constraints.

Additionally, in this study, one more idea was attempted for the DRL framework: positional encoding. While the invariant strategy aims at policy generalization and forcing the embedding of the structure of the equations into the policy and critic networks, the positional-encoding strategy aims at policy specialization. As shown in **Fig. 3(b)**, if perfect translation and symmetry-invariant actions should be taken regardless of the agent's position, all four segments should take the same action. The only way to take the same action while keeping the mean temperature constant is to keep the temperature change of all segments to zero. It means that the perfect invariant action imposes constraints on discovering efficient control trajectories. Even in a symmetrical flow field, other actions are often required depending on the segment location for efficient control. This conflict of agents is more likely to occur in a spatially periodic flow, such as this RBC phenomenon. It was noted that the same agent conflict problem can occur if each agent is affected by different boundary conditions, even in a chaotic flow. To resolve the excessive constraints of invariants, the state must include the location information of the segment. In this case, we can provide information about actuator spatial location to the network using the transformer model and leveraging positional encoding [31]. As a result, we develop a novel MARL architecture incorporating positional encoding, as shown in **Fig. 2(e)**. More detailed descriptions of positional-encoding-embedded DRL are provided in *Section 5*.

## 4. Invariant representation – group-invariant neural networks

### 4.1 Group-equivariant neural networks

Convolution correlation, widely known in signal processing as the measure of overlap between two signals when one is reflected and shifted, is commonly used in CNNs for image processing [41, 42]. As shown in **Eq. (7)**, performing the kernel operation as convolutional correlation allows us to capture the locality of pixel dependencies and the stationarity of statistics from image data. Here, $\underline{f} = (f_1, \ldots, f_N)$ represents the feature maps, $\underline{k} = (k_1, \ldots, k_N)$ represents the convolutional kernels, $N$ is the number of channels in input, and $\mathcal{T}_{\mathbf{x}}$ is the translation operator for translations $\mathbf{x}$ where $\mathbf{x}, \mathbf{x}' \in \mathbb{R}^2$. Because translational equivariance is guaranteed in the convolutional layer, CNNs that include fully connected layers can achieve high performance in image classification [43] and many more tasks that consider data with translational equivariance or invariance. However, unlike translation, convolution correlation does not guarantee rotation or symmetry equivariance. As an illustration of this fact, it has been observed that the performance of CNNs decreases when the images to be classified are rotated [44].

$$\left(\underline{k} \otimes_{\mathbb{R}^2} \underline{f}\right)(\mathbf{x}) \coloneqq \left(\mathcal{T}_{\mathbf{x}}\underline{k}, \underline{f}\right)_{\left(\mathbb{L}_2(\mathbb{R}^2)\right)^N} = \sum_{c=1}^{N} \int_{\mathbb{R}^2} k_c(\mathbf{x}' - \mathbf{x}) f_c(\mathbf{x}') \mathrm{d}\mathbf{x}', \qquad (7)$$

To address the problem of rotating images, Lafarge et al. developed roto-translation equivariant networks, known as group convolutional neural networks (G-CNNs) [30]. They developed convolutional correlation from translation group to roto-translation group based on group theory. G-CNNs are capable of performing equivariant operations under both translation and rotation conditions. Understanding the working principles of G-CNNs involves comprehending three types of layers: lifting, group convolutional, and pooling layers. First, in lifting layers, the input image undergoes a convolutional operation with a set of rotating kernels [30]. This operation is defined as follows:

$$\left(\underline{k} \,\widetilde{\otimes}\, \underline{f}\right)(g) \coloneqq \left(\mathcal{U}_g\underline{k}, \underline{f}\right)_{\left(\mathbb{L}_2(\mathbb{R}^2)\right)^N} = \sum_{c=1}^{N} \int_{\mathbb{R}^2} k_c(\mathbf{R}_\theta^{-1}(\mathbf{x}' - \mathbf{x})) f_c(\mathbf{x}') \mathrm{d}\mathbf{x}', \qquad (8)$$

where $\mathcal{U}_g$ is the operator for both planar translations (in $\mathbb{R}^2$) and rotations (in $SO(2)$). $\mathbf{R}_\theta$ is the rotational transformation. In other words, each rotated kernel ($on\ SO(2)$) is computed with the input image identically through convolutional correlation ($\mathbb{R}^2$). We can denote the group of roto-translations $SE(2) = \mathbb{R}^2 \rtimes SO(2)$. In **Fig. 4(a)**, for simplicity, the illustration was drawn on a symmetrical equivariance rather than a rotational one. An example of G-CNN operation in the $SE(2)$ rotation dimension is depicted in Ref. [30]. The output of the lifting layer is roto-translationally equivariant.

Images processed in the lifting layer ($SE(2)$ dimension) are subsequently processed by group convolutional layers, which perform a convolutional operation with a kernel set composed of the $SE(2)$ dimension as shown in Eq. (9). $\mathcal{L}_g$ is an operator similar to $\mathcal{U}_g$ but operates on a set of $SE(2)$ kernels $\underline{K}$ where $g = (x, \theta) \in SE(2)$. The Input is also the $SE(2)$ feature maps $\underline{F}$. As illustrated in **Fig. 4(a)**, unlike the lifting layer where there is only one set of rotated kernels, the group convolutional layer has multiple sets of rotated kernels that are shifted as shown in **Eq. (10)** [30]. $N'$ is the number of channels in the feature maps.

$$(\underline{K} \otimes \underline{F})(g) := \sum_{c=1}^{N'} (\mathcal{L}_g K_c, F_c)_{\mathbb{L}_2(SE(2))} = \sum_{c=1}^{N'} \int_{SE(2)} K_c(g^{-1} \cdot g') F_c(g') dg', \tag{9}$$

$$(\mathcal{L}_g F)(g') = F(g^{-1} \cdot g') = F(\mathbf{R}_\theta^{-1}(\mathbf{x}' - \mathbf{x}), \theta' - \theta), \tag{10}$$

Consequently, the lifting and group convolutional layers are equivariant with respect to both rotation and translation transformation, as follows:

$$\Phi(\mathcal{R}_g(f)) = \mathcal{R}'_g(\Phi(f)) \tag{11}$$

Where $\Phi$ is an operator $\mathbb{L}_2(X) \to \mathbb{L}_2(Y)$ by lifting layer and group convolutional layer, respectively. $\mathcal{R}_g$ and $\mathcal{R}'_g$ representations of G on respectively functions the domain $X$ and $Y$ [30]. Finally, the projection layer of G-CNNs imposes roto-translation invariance on neural networks while transforming an input $SE(2)$-image onto $\mathbb{R}^2$ [30]. Lafarge et al. noted and empirically showed that maintaining roto-translation equivariance through the lifting and group convolutional layers (before the pooling layer) can increase classification performance in histopathology image analysis applications [30]. This is because invariant operations can lose information too early in the context of deep learning, thereby reducing the network's ability to learn locality. By preserving equivariance longer, the network retains more detailed information, enhancing its overall performance [45].

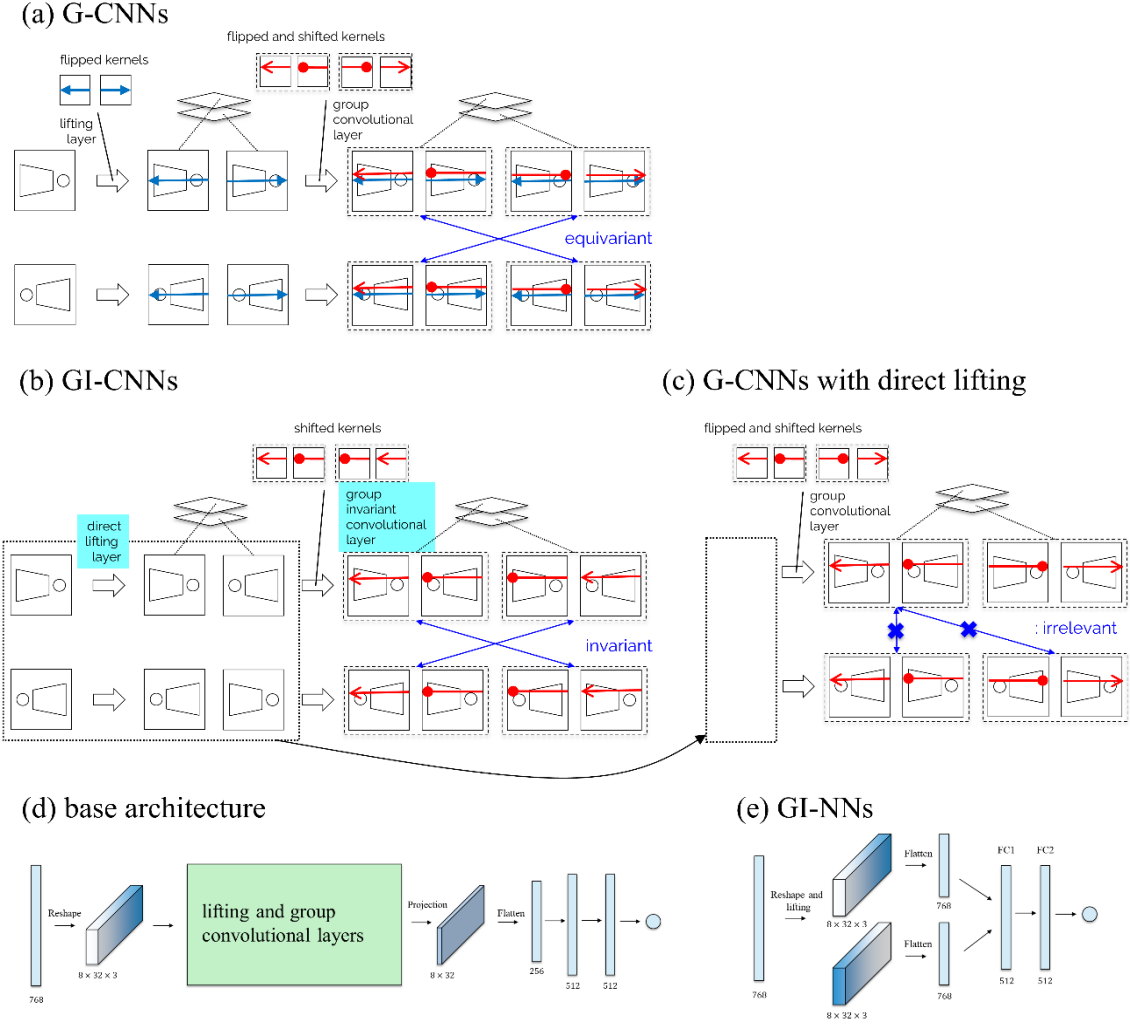

**Fig. 4.** Implementation methodology of (a) G-CNN, (b) GI-CNN and (c) G-CNNs with direct lifting based on a symmetric image example. It was noted that the direct lifting layer alone cannot be invariant. The lifting and group convolutional layers are implemented in the based architecture's green box in (d). For symmetrical invariance, kernels are only shifted (not flipped) in GI-CNNs. (E) Full network architecture of GI-NNs.

### 4.2 Development of group-invariant convolutional neural networks

As discussed in *Section 3*, our flow-control problem requires a sym-translation invariant policy for the flow-field state, as shown in **Fig. 3(a)**. Hereafter, we refer to symmetrical translation as 'sym-translation'. This requirement arises because identical temperature control is to be expected when the flow field is symmetric relative to the agent's position. With G-CNN's kernel operation method, it is challenging to calculate the same action when symmetric images are input. Although maintaining equivariance in the early layers of the deep neural network can improve classification performance, the characteristics of RL in flow-control problems are different from those of classification. First, the output dimension of deep neural networks in RL is not binary but spans a wide continuous action space. In a continuous output condition, it is difficult to make equivariant feature maps invariant simply by using a later fully connected layer or a pooling layer. Second, unlike supervised learning, which can leverage sufficient image data for training, RL relies on data generated through interaction with the environment at each timestep, making it challenging to empirically learn invariants. This implies that the RL performance can increase when the architecture of the network itself can produce invariant output values for the targeted transformation. This is one of the reasons why MARL has been successful in previous studies on translational transformation [22]. Because G-CNNs themselves cannot produce sym-translation invariant output by network architecture, a sym-translation invariant network was developed in this study to further improve DRL performance.

As shown in **Fig. 4(b)**, we developed group-invariant convolutional neural networks (GI-CNNs) with a novel network architecture for symmetric invariant policy. To describe GI-CNNs using the same nomenclature as G-CNNs, **Eq. (12)** first shows the direct lifting layer with respect to rotation transformation, $U_\theta$. The direct lifting layer transforms an input $R^2$-image onto $SE(2)$ feature maps $\underline{F}$ without kernel operation. In other words, the input image $\underline{f}$ is transformed into a set of all rotated images $\underline{F}$ as depicted in **Fig. 4(b)**. The number of feature map $N$ is 3 (temperature, $x$-axis velocity, $y$-axis velocity). This direct lifting operator is completely invariant with respect to rotational transformations. Although it is not invariant regarding translation transformation, our MARL framework is already translation invariant at the segment level. This explains why the base network architecture is not translation invariant without MARL implementation.

$$\underline{F} \coloneqq \left(\mathcal{U}_\theta \underline{f}\right)_{\left(\mathbb{L}_2(\mathbb{R}^2)\right)^N} = \underline{f}(\mathrm{R}_\theta^{-1}) \tag{12}$$

For the purpose of this study, the direct lifting layer performs a symmetric transformation rather than a rotation transformation as shown in **Eq. (13)**. This is represented by $F$, the flip operation. In this symmetric transformation, there are only two pairs of feature map bundles as shown in **Fig. 4(b)**. In rotational transformation, the number of bundles is determined by the angle hyperparameter [30].

$$\underline{F} \coloneqq \left(\mathcal{U}_\mathcal{F} \underline{f}\right)_{\left(\mathbb{L}_2(\mathbb{R}^2)\right)^N} = \underline{f}(\mathcal{F}^{-1}) \tag{13}$$

Following the direct lifting layer, operations are performed in the group invariant convolutional layer. Using the same equation format as the group convolutional layer (**Eq. (9)**), the group invariant convolutional layer is defined as follows:

$$(\underline{K} \otimes \underline{F})(g) := \sum_{c=1}^{N} (\mathcal{J}_g K_c, F_c)_{\mathbb{L}_2(SE(2))} = \sum_{c=1}^{N} \int_{SE(2)} K_c((\mathbf{x}' - \mathbf{x}), \theta' - \theta) F_c(g') dg' \quad (14)$$

The main difference is that in the $\mathcal{J}_g$ operator, the kernel is only shifted and not rotated, as shown in **Eq. (15)** and in the group-invariant convolutional layer in **Fig. 4(b)**. This $SE(2)$-kernel operation method is structurally identical to the group convolutional layer, although it is convolutional only for the translation group. Therefore, we name it the group-invariant convolutional layer. This characteristic allows GI-CNNs to be invariant to rotational transformations. If a group convolutional layer was adopted, the output would not be invariant with respect to rotation, as illustrated in **Fig. 4(c)**. The group-invariant convolutional layer in the sym-translational invariant network used in this study is shown in **Eq. (16)**. The number of kernels used in the group-invariant convolutional layer is 1024 (3× 3 window size). It is noted that much smaller parameters were used (420,864) than the base FC (655,360)**.**

$$(\mathcal{J}_g F)(g') = F((\mathbf{x}' - \mathbf{x}), \theta' - \theta) \quad (15)$$

$$(\mathcal{J}_{g_\mathcal{F}} F)(g_\mathcal{F}') = F((\mathbf{x}' - \mathbf{x}), \mathcal{F}) \quad (16)$$

Consequently, the MARL-based group-invariant convolutional network is not equivariant (**Eq. (17)**), but invariant with respect to roto-translation transformation as shown in **Eq. (18)**.

$$\Phi(\mathcal{R}_g(f)) \neq \mathcal{R}'_g(\Phi(f)) \quad (17)$$

$$\Phi(\mathcal{R}_g(f)) = \Phi(f) \quad (18)$$

**Fig. 4(d)** shows the base architecture for the DRL agent. The input flow and temperature fields are reshaped into the original image format. The DRL algorithm with MARL-applied GI-CNNs can be expressed in the Algorithm 2:

```
Algorithm 2 PPO-Clip in invariant representation MARL framework
Input: initial policy parameters θ₀, initial value function parameters ϕ₀
for k = 0, M do
  for t = 0, T do
    Execute merged action aₜ
      for i = 0, N do
        Recentering s_{i,t} ← s'_{i,t}
        Compute advantage estimates Â_{i,t} with recentered state s_{i,t} using GI-CNNs
        Compute action a_{i,t} following π_θ with recentered state s_{i,t} using GI-CNNs
    end for
  end for
  Optimize surrogate L wrt θ, ϕ
```

$$\theta_{k+1} = \underset{\theta}{\arg\max} \frac{1}{|D_k|T} \sum_{\tau \in D_k} \sum_{t=0}^{T} \sum_{i=0}^{N} \min\left( \frac{\pi_\theta(a_{i,t}|s_{i,t})}{\pi_{\theta_k}(a_{i,t}|s_{i,t})} A^{\pi_{\theta_k}}(s_{i,t}, a_{i,t}), g(\epsilon, A^{\pi_{\theta_k}}(s_{i,t}, a_{i,t})) \right)$$

$$\phi_{k+1} = \underset{\phi}{\arg\min} \frac{1}{|D_k|T} \sum_{\tau \in D_k} \sum_{t=0}^{T} \sum_{i=0}^{N} (V_\phi(s_{i,t}) - \hat{R}_{i,t})^2$$

```
  via gradient descent algorithm
end for
```

Global translational invariance is imposed through MARL's recentering operation, and local symmetrical invariance is imposed through GI-CNNs. The difference between global and local is whether the invariance imposed by transformation is segment-wise or spatial continuous. The actor network and critic network are composed of the same GI-CNN architecture. Observing performance changes when configuring the two networks differently is part of our future plan.

### 4.3 Effects of symmetric invariant representation

Due to their novel network architecture and kernel operation, GI-CNNs can guarantee that their outputs are invariant under symmetric (or rotational) input images. When the action for one specific flow field $T, u, v$ is updated, the action for the symmetric field is also updated. If our target system is completely action-invariant with respect to symmetry, we can reduce the state space required for training by half. This reduction in state space increases the ratio of the state covered in one episode to the total state, thereby enhancing policy convergence speed. In this study, the number of action steps per episode is 200, with each action lasting 1.5 time units, resulting in an episode duration of 300 time units. For the remaining parameters in DRL-RBC framework, we use the same values used in [22]. Note that in the current study, the parameters not related to feature representation are set to the same values as in [Ref.] to focus on the effects of the representation.

We analyze the effects of symmetrical invariant representation by comparing the learning curve of the base MARL and our developed architecture (GI-CNNs applied MARL). The base MARL uses a fully connected layer, which has 512 units per hidden layer and 2 hidden layers [22]. To mitigate the potential bias from a single learning curve, each training run is performed twice. The solid and dotted lines represent the average learning curve and each individual learning curve, respectively. The curves are shown based on the moving average for 25 episodes and the fluctuations of each episode are also illustrated. Additionally, to prevent misunderstanding of the DRL performance, if unlearning occurs even in a single run, the moving average is not calculated.

The black and green lines in **Fig. 5** represent the learning curves of the base MARL and of our method, respectively. As expected, the convergence speed of GI-CNNs improves compared to the base MARL. In the base MARL, the average *Nu* converges after 250-300 episodes, which is identical to the observations made in previous study [22]. Even when analyzing single runs, the number of required episodes always exceeds 250. In contrast, the average *Nu* of GI-CNNs is achieved after around 150-200 episodes. Although *Nu* tends to increase slightly again, there are no steep fluctuations as seen in the base architecture. However, contrary to our expectations, the performance of the GI-CNNs is diminished compared to the base MARL. We observe that the minimum *Nu* achieved through control is greater than that of base MARL. We suspected that one of two factors in GI-CNNs was responsible for the poor performance: 1) noise from zero padding, or 2) losing equivariance too early

During convolution operations, we used the zero-padding technique to prevent boundary data loss in GI-CNNs. Although the CNN is an effective approach for extracting stationarity of statistics from image formats, zero padding can reduce accuracy in regression tasks. The noise introduced by zero padding becomes more significant, especially as the image size decreases. It should be noted that the state in our DRL framework is a small 8×32 image format. Apart from the global translational invariance achieved through the multi-agent approach, the negative effect of GI-CNNs losing translational equivariance early may be more significant. To investigate which of the two factors is the main cause, we develop fully connected (FC)-based group-invariant neural networks (GI-NNs) in the next section.

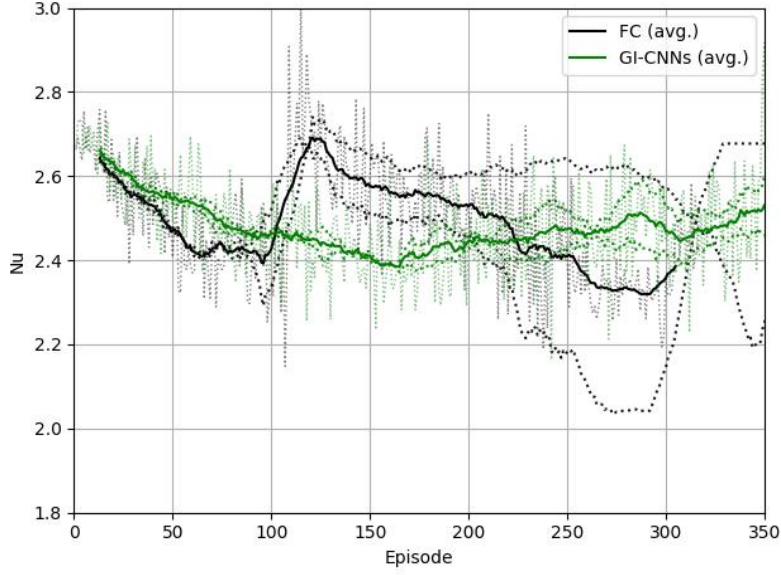

**Fig. 5.** Effects of symmetric invariant representation investigated by GI-CNNs. The solid and dotted lines are moving $Nu$ averages for 25 episodes in multi-runs (averaged) and single-run, respectively. Additionally, the transparent dotted lines are instantaneous $Nu$ based on the averaged multi-runs. If unlearning occurs even in one single run, the moving average is not calculated.

### 4.4 Group-invariant fully connected neural networks

To exclude noise caused by zero padding, we develop group-invariant neural networks with fully connected layers (GI-NNs), as shown in **Fig. 4(e)**. In GI-NNs, the state is symmetrically transformed and then individually processed by common fully connected layers. The two resulting values are finally combined to determine the action. With this architecture, the same action is produced when the symmetric state is an input to the neural networks, as when the original state is an input. This symmetric-invariant property can be understood mathematically in **Eqs. (19) and (20)**. Here $a$ is an input feature, $a_F$ is a flipped input feature by using the reshape and flatten functions, $W$ is the bias and $\hat{b}$ is the output. This structure ensures that the network produces invariant outputs for symmetric inputs, thereby eliminating the noise introduced by zero padding in CNNs and maintaining the desired symmetry in the actions.

$$\hat{b} = \sigma(aW_1)W_2 + \sigma(a_F W_1)W_2 \tag{19}$$

$$\hat{b} = \sigma(aW_1)W_2 \tag{20}$$

Because the activation-function calculation is performed before the values are combined, the averaging problem is solved. In other words, if the architecture shares the activation function for $a$ and $a_F$, the output of $a$ becomes identical to the output of $\frac{a+a_F}{2}$ not only $a_F$ as shown **Eq. (21)**.

$$\hat{b} = \sigma((a + a_F)W_1)W_2 \rightarrow \hat{y}(a) = \hat{y}\left(\frac{a+a_F}{2}\right) \tag{21}$$

**Fig. 6** compares the performance of DRL architectures under the same conditions as in **Fig. 5**. Similar to GI-CNNs, GI-NNs converge more efficiently than the base MARL because they explore an optimal strategy in the symmetry-reduced subspace. The average $Nu$ of GI-NNs converges after around 200–250 episodes. It is noted that, in addition to the convergence speed, the average performance of the policies is also significantly improved. This suggests that the main reason for the inferior performance of GI-CNNs is the noise from zero padding in the small image conditions. Furthermore, the learning curves are very similar until unlearning occurs in both runs (blue dotted line). This increased learning reproducibility of RL is consistent with the results from the previous GI-CNNs. In other words, as the complexity of the state decreases, the agent follows a similar learning process even in independent runs. This increase in reproducibility is highly valuable in RL, where computational cost is a significant burden. RL frameworks with low reproducibility must find the optimal policy through repeated runs, which is computationally expensive. By enhancing reproducibility, GI-NNs reduce the need for multiple runs, making the RL process more efficient and cost-effective.

When we compare the results based on each run, the minimum $Nu$ value is observed in the base MARL learning curve (black star). The minimum $Nu$ in the base MARL is about 0.05 smaller than in GI-NNs. Differences of this magnitude can be considered as stochastic effects in the action decision process rather than a difference in optimal policy quality. We analyze the RBC control process to further improve the quality of the optimal policy. **Fig. 3(b)** shows the positions of each agent in the initial conditions of RBC control. It is important to note that in our multi-agent based GI-NNs, not only translational but also symmetry-invariant properties are imposed on the agent. In other words, the same action is determined in states that are translationally or symmetrically equivalent based on the agent's location. When analyzed based on the initial condition, we see that the agents have constraints that make it difficult to take different actions that perform symmetry breakings that may be necessary to obtain better flow efficiency. Even in an invariant state, it is disadvantageous in situations where it is more effective to take different actions depending on the agent's location. Because it is easy to satisfy translational or symmetry conditions, this action constraint problem is more prominent in a spatially periodic flow like RBC. Said otherwise, strictly imposing invariance and symmetry of the policy is too restrictive, as it prevents the DRL agent from being able to break such symmetry, even in cases when this is actually beneficial.

We conclude that investigating the effect of unique representation, where each agent considers its own location when deciding on an action (by contrast to a purely invariant representation), is also essential for efficient DRL architecture development. This approach could mitigate the constraints imposed by invariance and potentially lead to better performance in situations requiring location-specific actions.

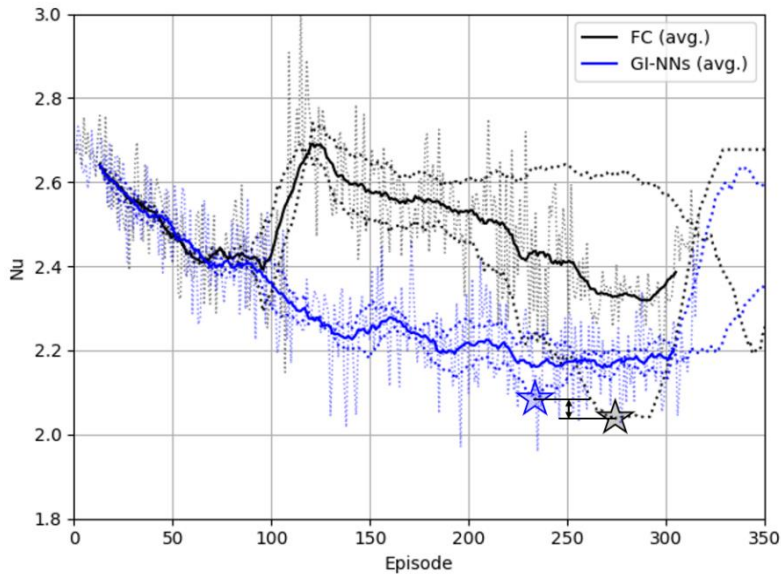

**Fig. 6.** Effects of symmetrical invariant representation by GI-NNs. It is confirmed that both learning speed and quality (*Nu* minimization) are improved over base MARL. Note that learning reproducibility is greatly enhanced in GI-NNs.

## 5. Unique representation – positional encoding

### 5.1 Concept and implementation of positional encoding

Recently, transformers have emerged as a strong alternative to CNNs and RNNs in visual and time-series tasks [31, 46, 47]. Compared to the former LSTM networks for sequence processing, the self-attention mechanism of the transformer improves the ability to process lengthy sequences [31]. Although the self-attention mechanism can dynamically adapt the scope of focus depending on the input data region, it is permutation-invariant, which discards the order of the input sequence. To address this issue, the positional-encoding method was proposed in previous works [47]. By adding absolute positional encoding to each input token, order-awareness is enabled, significantly enhancing the performance of the transformer. Ref. [47] illustrates how positional encoding is implemented in the typical transformer architecture. Positional encoding is added to the input embedding at the bottoms of the encoder and decoder stacks. We rewrite the sine and cosine functions with different frequencies that were used in [47] as shown below:

$$PE_{pos,2i} = sin\ (pos/1000^{2i/d_{model}}), \qquad (22)$$

$$PE_{pos,2i+1} = cos\ (pos/1000^{2i/d_{model}}) \qquad (23)$$

Where $pos$ is a token position, $d_{model}$ is the size of token embedding vector. Interestingly, the motivation behind this approach is identical to that of this study, i.e. it aims to resolve network performance degradation caused by invariance. The next section will demonstrate how positional encoding can be embedded into our DRL framework. We now describe the manner in which positional encoding is implemented within the MARL framework used in this study. **Fig. 2(e)** and **Algorithm 3** show our MARL framework with the implementation of positional encoding. Before the variable fields are recentered for single agent learning in the MARL framework, a positional encoding matrix is added to the temperature field. This positional encoding indirectly provides absolute positional information to the agent. To prevent overfitting to positional information, positional encoding is embedded only in the temperature variable. In the transformer model, sine and cosine functions are used together to distinguish both the location and dimension of the token. However, since we only need to distinguish the $x$-position of the agent in the variable fields, a one-dimensional sine function is used here as shown in **Eq. (24)**. $x_{probes}$ represents the $x$-coordinate of each probe point. The probe points at both ends have $s_{PE}$ values of 0.

$$s_{PE}(x,y) = sin\ (\frac{x_{probes}}{2\pi}) \qquad (24)$$

It is important that the variable field with added positional encoding is used only to determine the RL's action, and the reward calculation remains unchanged. By embedding positional encoding in this manner, we enhance the network's ability to leverage spatial information effectively without compromising the learning process.

**Algorithm 3** PPO-Clip in unique representation MARL framework

Input: initial policy parameters $\theta_0$, initial value function parameters $\phi_0$
**for** $k = 0, M$ **do**
  **for** $t = 0, T$ **do**
    Execute merged action $a_t$
      **for** $i = 0, N$ **do**
        Positional encoding $s_{i,t} \leftarrow s_{i,t} + s_{PE}$
        Recentering $s_{i,t} \leftarrow s'_{i,t}$
        Compute advantage estimates $\hat{A}_{i,t}$ with recentered state $s_{i,t}$
        Compute action $a_{i,t}$ following $\pi_\theta$ with recentered state $s_{i,t}$
    **end for**
  **end for**
Optimize surrogate $L$ wrt $\theta, \phi$

$$\theta_{k+1} = \underset{\theta}{\operatorname{argmax}} \frac{1}{|D_k|T} \sum_{\tau \in D_k} \sum_{t=0}^{T} \sum_{i=0}^{N} \min\left(\frac{\pi_\theta(a_{i,t}|s_{i,t})}{\pi_{\theta_k}(a_{i,t}|s_{i,t})} A^{\pi_{\theta_k}}(s_{i,t}, a_{i,t}), g\left(\epsilon, A^{\pi_{\theta_k}}(s_{i,t}, a_{i,t})\right)\right)$$

$$\phi_{k+1} = \underset{\phi}{\operatorname{argmin}} \frac{1}{|D_k|T} \sum_{\tau \in D_k} \sum_{t=0}^{T} \sum_{i=0}^{N} \left(V_\phi(s_{i,t}) - \hat{R}_{i,t}\right)^2$$

via gradient descent algorithm
**end for**

## 5.2 Effects of positional encoding

**Fig. 7** compares the DRL performance between the FC (base MARL) and positional encoding embedded FC (PE-FC) cases. In a single run, the PE-FC architecture, which provides agent location information to the MARL framework, significantly reduces $Nu$ even from the first episode. Due to the stochastic action procedure, another run shows a very similar learning curve until about the 80th episode. In FC, a region where $Nu$ rapidly increases as the learning information is initialized is commonly observed. As in [22], this learning memory loss occurs more clearly in MARL. Because the same neural network parameters are shared between agents responsible for each segment, parameter updates between agents may conflict. The constraints on these actions become even greater in group invariant networks where symmetrical invariants are also imposed. On the other hand, the PE-FC architecture continues to decrease $Nu$ without a noticeable increase region, although there are some fluctuations due to the random exploration nature of RL. When different actions are needed in the same state (recentered state) depending on the agent's location (in the MARL framework), the base FC cannot calculate these different actions from the same parameters and inputs. By contrast, PE-FC can obtain these different actions because location information is embedded in the temperature field. Due to these positional encoding effects, PE-FC reaches the minimum $Nu$ of the FC case in less than half of the episodes and shows a stable convergence pattern. This can cut RL training time in half, a very encouraging savings for flow control problems involving expensive numerical simulations.

Because the improved learning curves are confirmed, we conclude that the positional encoding has a negligible negative impact on undermining the invariant purpose of MARL. Our conclusion aligns with the observation that embedding positional encoding significantly increases network performance, even if it introduces some noise in the attention scores in transformers. The potential impairment of generalization performance due to PE can ultimately be overcome with increased training data. In this case, optimal $Nu$ reduction and generalization performance by PE-FC can be targeted simultaneously. In this study, the weighting factor of the sine function positional encoding was set to unity. Optimizing generalization and specialization according to changes in this weighting factor is let for future work.

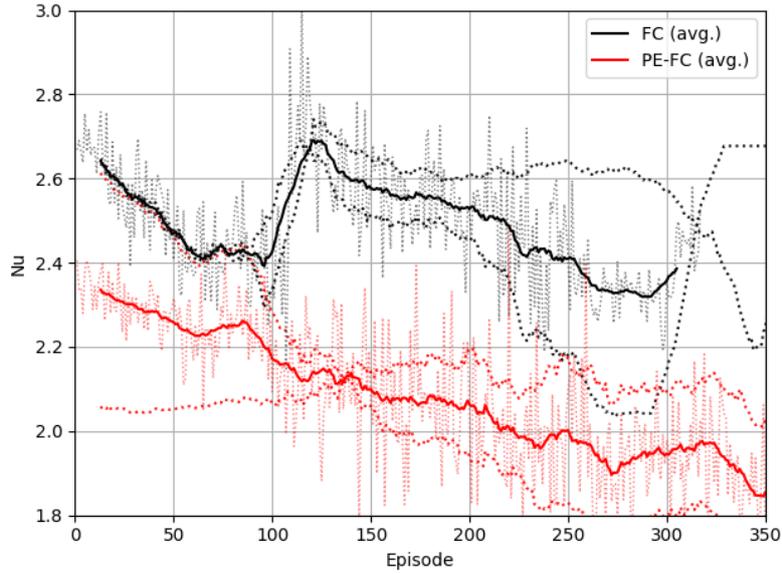

**Fig. 7.** Effects of unique representation on learning curve demonstrated by PE-FC. Notably, PE-FC as proposed for the first time in this study shows the best performance among all attempted models. More explanation We should note that the combination of invariant and unique representation can effectively improve the DRL performance, depending on the characteristics of the target control problem.

### 5.3 Verification of PE-FC method

It has, therefore, been confirmed that the MARL framework applying PE-FC significantly improves performance compared to the base MARL. In particular, the number of episodes for optimal policy convergence is noticeably reduced. However, it is necessary to verify in detail the RBC phenomenon during the control process to determine whether the control strategy is physically feasible. In this section, the behavior of $Nu$ within an episode, rather than the average of the final converged $Nu$ values of the episodes, isinvestigated both quantitatively and phenomenologically. The DRL framework trained through episodes has fixed neural network parameters. In other words, the optimal policy that can calculate the best action for each state is stored in the neural network. Because the PPO algorithm belongs to the stochastic policy gradient method, the mean and variance of the action are calculated for each state, and hence the action is determined stochastically $a \sim \pi_\theta(.|s)$ [35, 36]. The dotted line in **Fig. 8** shows how the trained PE-FC (stochastic mode) controls the $Nu$ of RBC in a single episode. The $x$-axis represents the actuation (action). Although the DRL framework has been trained, $Nu$ still shows large fluctuations in the stochastic run. This is because the distribution of actions is still wide, as can be seen from the action variation in each segment in **Fig. B1** where the actions are plotted based on normalized values. Due to this randomness, the behavior of $Nu$ can vary depending on each run.

Therefore, as shown by the solid line in **Fig. 8**, the performance of the trained DRL framework is evaluated deterministically (deterministic mode). In a deterministic mode, only the action with the highest probability is selected, $a = \pi(s)$. Consequently, if the initial conditions are the same, the $\boldsymbol{Nu}$ behavior in each run is almost identical. Additionally, because there is no chance that a bad action will be selected through the probability distribution, it is confirmed that the system converges more quickly to the lowest $\boldsymbol{Nu}$, and fluctuations virtually disappear. It is noted that the deterministic runs of the base FC took about 200 actuations for $\boldsymbol{Nu}$ to completely converge [22], while PE-FC converges in just 100 actuations. Because the action conflict issue between agents is resolved, RBC flow can be controlled through a shorter path. **Fig. B1** shows the action variation in this deterministic run, clearly identifying that the action variation is smooth in all segments. As a result, it is confirmed that positional encoding not only increases learning efficiency but also enhances the deterministic performance of the network after training.

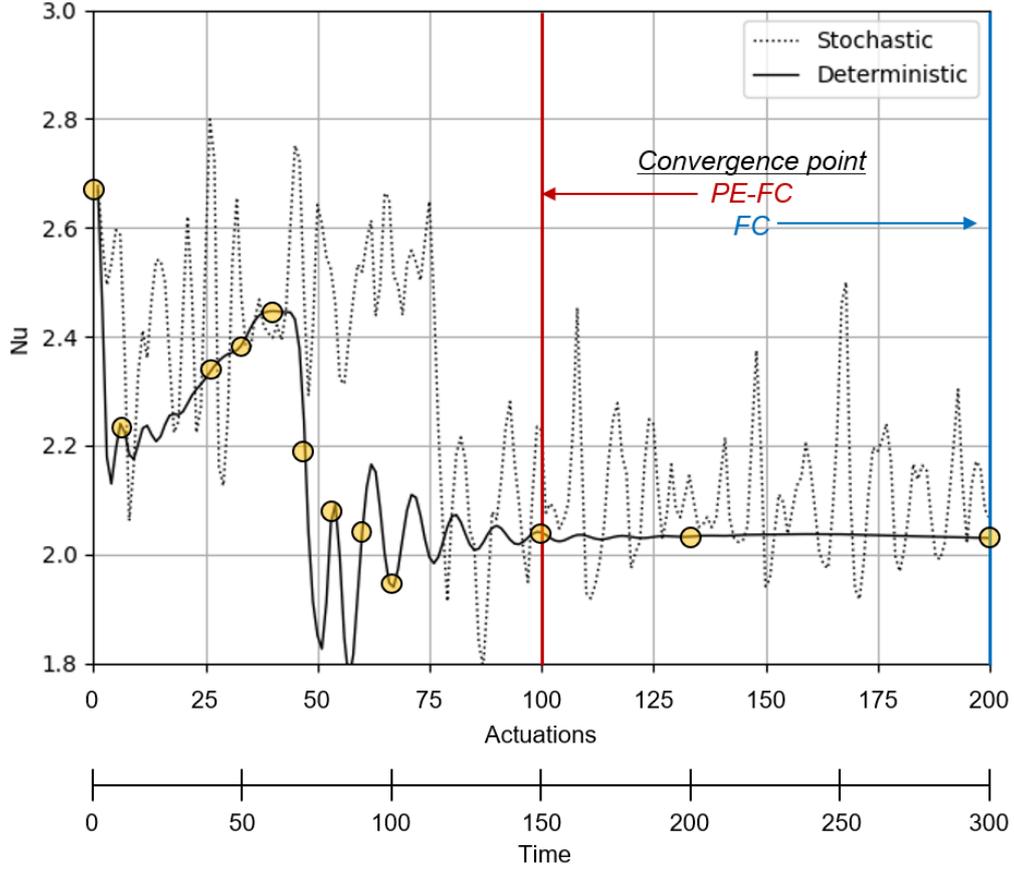

**Fig. 8**. Comparison of stochastic and deterministic run of trained PE-FC model. Unlike deterministic, stochastic executes actuation as probability distributions sampling following the calculated mean and variance, which is the key factor that enables exploration to take place. It should be noted that *Nu* convergence in deterministic run is 100 actuations faster than obtained with base MARL in this and previous studies.

To phenomenologically investigate the validity of the trained PE-FC model's policy, we analyze the sequence of temperature and velocity fields. **Fig. 9** illustrates the temporal evolution of temperature and velocity fields with the 10 control segments. Key points and features over time include the formation and evolution of convection cells. At 1 time units, heat control begins in the flow field of the baseline. It should be noted that heat has been concentrated in segments 1, 5, and 10 since then. After experiencing many episodes during training, the agent learned that merging two convection cells can effectively reduce $\boldsymbol{Nu}$. A straightforward strategy for merging two cells is to suppress the downward flow along the outer periphery of each cell to form a combined uplift region. For this reason, in segments 1 and 10, high wall temperatures continue to be imposed as actions. Also, heat is concentrated in segment 5 because the central downward flow must be shifted to the right to match the combined single convection cell location. The upward momentum generated by buoyancy in segment 5 combines with the downward flow and flows to the left into the single convection cell. By this strategy, as time progresses (40-100 time units), the cells merge and form a more stable and larger 1 convection cell. Around 150-300 time units, the convection cells stabilize further, maintaining a consistent pattern. The system reaches a steady state with well-defined convection cells, indicating a stable temperature distribution and velocity field.

As shown in the 1 time units contour, the centered flow fields based on segments 1 and 6 are very similar. However, in the optimal policy, segment 1 requires an action with a temperature higher than segment 6. The MARL agent with translational invariance requires many episodes of training to carefully distinguish the minor, subtle differences between these two local flow fields with high-dimensional regression. If the agent cannot classify this difference, it will not be able to concentrate the temperature on segment 1, because it would have to give segments 5, 6, 10 the same temperature as 1. On the other hand, the PE-FC agent can more easily control segment 6 to a relatively lower temperature compared to segment 1. This is possible because positional encoding provides the agent with location information about the segment, thereby relaxing the constraints of invariance.

**Fig. 10** compares the actions and temporal fields between MARL and PE-MARL (PE-FC). It should be noted that the PE-MARL converged about 100 episodes faster but achieved exactly the same policy as MARL. As a result, positional encoding is a very effective representation technique when the agent is not perfectly translationally invariant or when tight invariant constraints or symmetry need to be broken to reach a better state. When the initial conditions change, the generalization performance of PE-MARL may decrease compared to MARL. Conversely, in more complex problems, the performance of MARL's optimal policy may decrease. Thus, a more rigorous investigation of PE-MARL and MARL is needed to balance optimal policy performance and generalization performance in a range of more sophisticated applications, which will be the focus of our future work. Additionally, Fig. B2 shows the temporal evolution of temperature and velocity fields in the stochastic run of the trained PE-FC model. Because the action is selected based on the probability distribution, it succeeds later in forming one convection cell compared to the deterministic run. This confirms the well known fact that stochastic policy gradient RL should generally be simulated as a deterministic run, unlike in the training phase, for final policy evaluation.

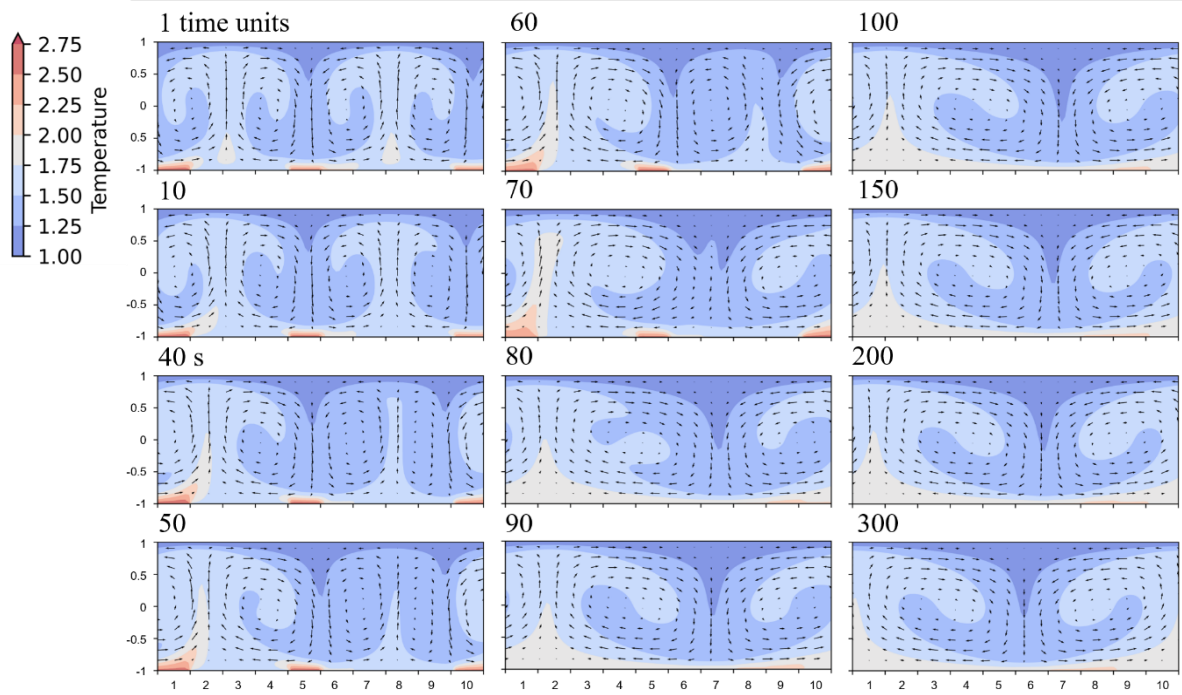

**Fig. 9.** Temperature contours from the deterministic run of trained PE-FC model at various time instants. Notably, by maintaining the heat concentrated segments for about 70 time units, the merged convection cells become stable in 100 time units. This is consistent with the approach of the optimal policy found in the base case, and proves that the optimal policy converges in pe-fc as well.

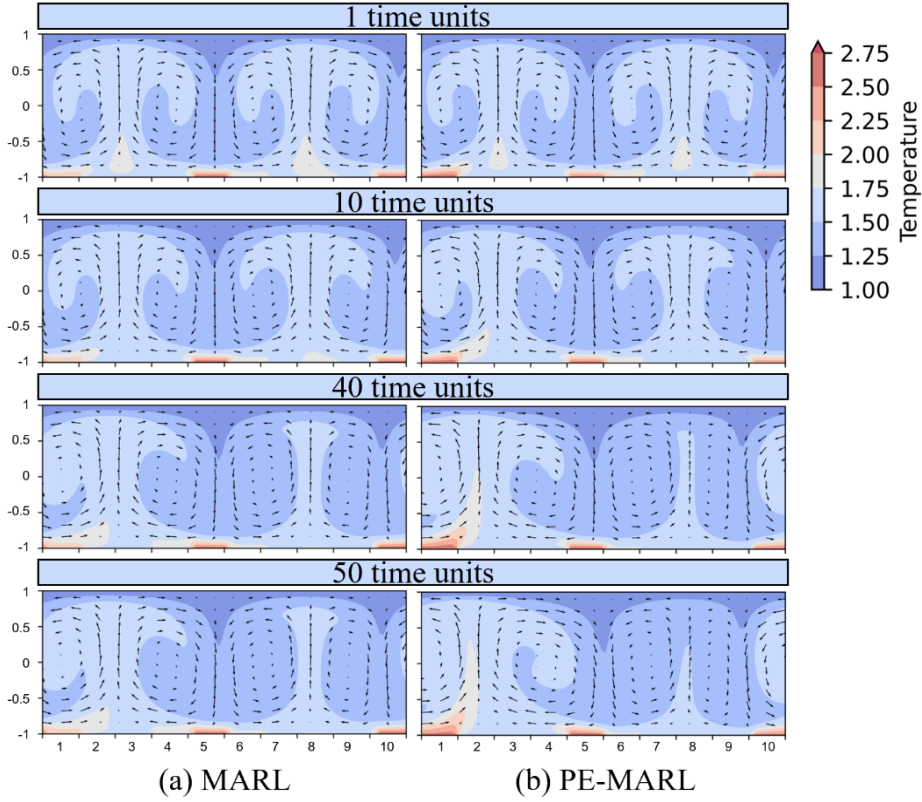

**Fig. 10.** Comparison of actions and temporal fields between MARL and PE-MARL. This confirms that PE-MARL converged to the same policy as MARL, even though the policy converged faster.

## 6. Conclusions

The proposed DRL methods integrate fluid-mechanics knowledge into DRL by developing advanced frameworks to address the complexity of flow-control systems. The novelty lies in combining group-invariant networks and positional encoding to improve DRL performance. Specifically, the study leverages MARL to exploit translational invariance and introduces a group-invariant network to exploit local symmetry invariance in the policy. Furthermore, it incorporates positional encoding inspired by transformer models to provide location information to agents, mitigating constraints resulting otherwise from excessive invariance. These innovations aim to enhance the efficiency of DRL in managing complex, high-dimensional flow-control problems. The performance evaluation focuses on the RBC problem, comparing the new DRL framework with previous methods. Our results demonstrate that both group invariant networks and positional encoding significantly improve learning efficiency and average policy performance. GI-NNs show faster convergence than the base architecture, achieving better average performance across episodes. Positional encoding further improves on these results by more effectively reducing the minimum $Nu$ and stabilizing convergence. The developed invariant and unique representation methods lead to robust and reproducible learning processes, validating the efficacy of the proposed methods for optimizing flow control in energy systems.

**Fig. 11** compares all DRL methods developed in this study with the baseline FC. The single-run results of the positional encoding embedded GI-NNs are also depicted as a purple line (PE-GI-NNs). Interestingly, the PE-GI-NNs still improve the quality of the optimal policy compared to GI-NNs. Consequently, group invariant networks are specialized in improving learning speed and positional encoding is specialized in improving learning quality. These results demonstrate that choosing a suitable feature-representation method according to the purpose as well as the characteristics of each control problem is essential. The results of this study are meaningful as they demonstrate that invariant and unique representation methods should be selected according to the characteristics of the target control problem. Moreover, we observe that a balance exists between not enforcing enough invariants and symmetries in the formulation of the DRL agent (in which case, the networks need to approximately re-learn the corresponding structures, resulting in increased learning cost and decreased performance), versus imposing these symmetries and invariants too rigidly (in which case it is challenging for the DRL agent to break symmetries in the system to control, even when this is advantageous). In future work, we plan to investigate the effects of boundary conditions and domain size on the efficacy of each method and investigate how to efficiently balance these two representation methods

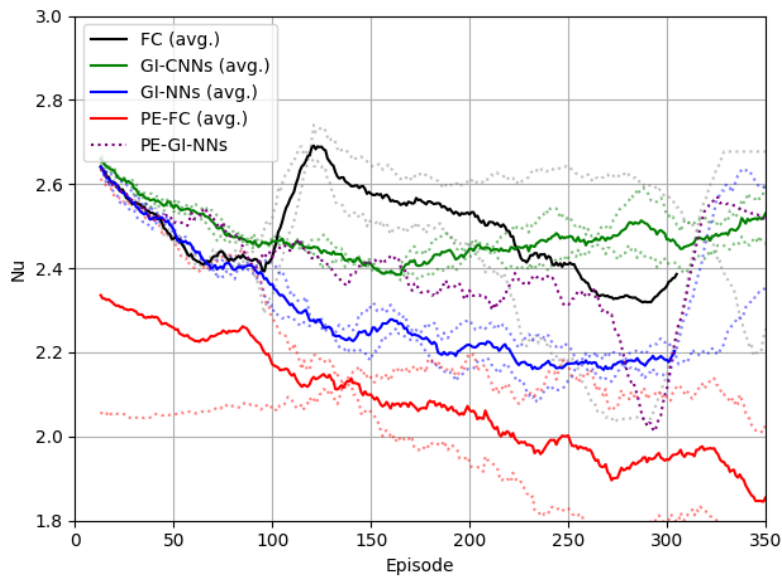

**Fig. 11**. Summary of DRL *Nu* minimization performance comparison according to DRL methods. We note the significant improvements achieved with our proposed methods. The solid and dotted lines are moving $Nu$ averages for 25 episodes in multi-runs (averaged) and single-run, respectively.

**ACKNOWLEDGEMENTS**

This paper was supported by research funds for newly appointed professors of Jeonbuk National University in 2023 and the "Human Resources Program in Energy Technology" of the Korea Institute of Energy Technology Evaluation and Planning (KETEP), which received financial resources from the Ministry of Trade, Industry &Energy, Republic of Korea (No. 20204010600470). RV acknowledges financial support from the ERC 764 Grant No. "2021-CoG-101043998, DEEPCONTROL". Views and opinions expressed are however those of the authors only and do not necessarily reflect those of the European Union or the European Research Council. Neither the European Union nor the granting authority can be held responsible for them.

**DATA AVAILABILITY**

The data that support the findings of this study may be generated from the code in the GitHub repository linked to this paper:

https://github.com/KTH-FlowAI/AdvancedDeepReinforcementLearning_RBC_Control

**AUTHOR DECLARATIONS**

The authors have no conflicts to disclose

**APPENDIX A: CODE RELEASE**

All the codes, scripts, and post-processing tools used in this work are made available on Github together with readmes and user instructions, at the following address: https://github.com/KTH-FlowAI/AdvancedDeepReinforcementLearning_RBC_Control [will be released openly over publication of this manuscript in the peer-reviewed literature]. Reasonable user support will be provided through the issue tracker of the corresponding Github repository.

**APPENDIX B: ACTION HISTORY**

   **Fig. B1** shows the comparison of action in the deterministic and stochastic run in this study. Compared to the stochastic run, the deterministic run's action variation is smooth in all segments. It is confirmed that positional encoding not only increases learning efficiency but also enhances the deterministic performance of the network after training. Additionally, **Fig. B2** shows the temporal evolution in the stochastic run of the trained PE-FC model. Because the action is selected based on the probability distribution, it succeeds later in forming one convection cell compared to the deterministic run.

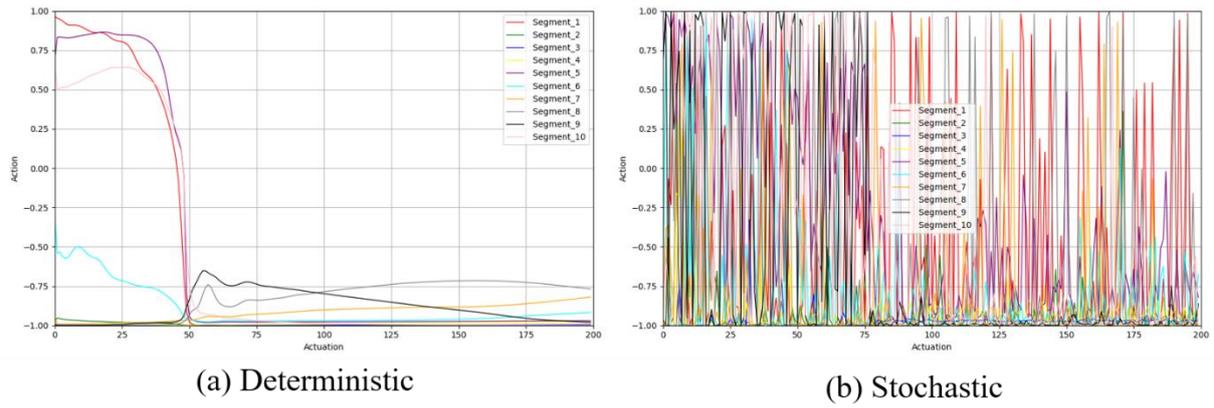

(a) Deterministic  (b) Stochastic

**Fig. B1**. Action history in each segment (agent) of the PE-FC's (a) deterministic run and (b) stochastic run. Because stochastic run involves exploration, variations show randomness. On the other hand, noise (exploration) from action is excluded, hence the deterministic run converges2 faster to minimum $Nu$.

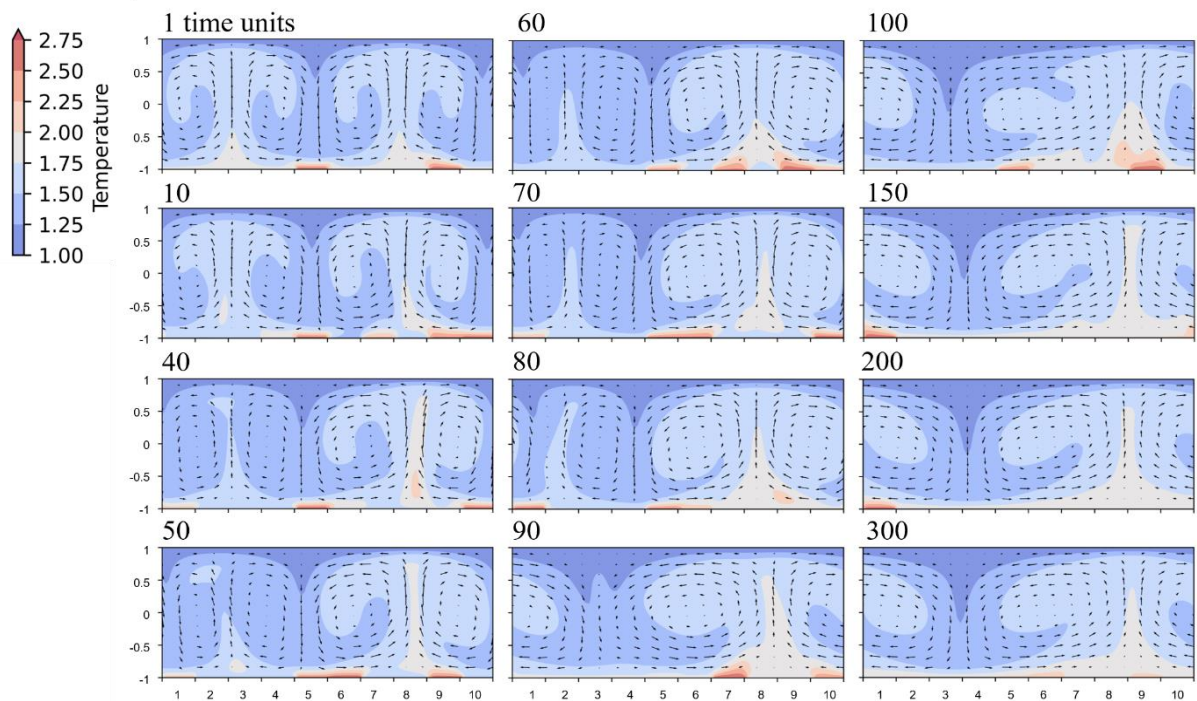

**Fig. B2.** Temporal evolution of temperature (contour) and velocity (arrows) fields in the stochastic run of trained PE-FC model. Because the action is actuated based on the probability distribution, it succeeds in forming one convection later than the deterministic run.

**References**

[1] J. Sanchez-Torrijos, K. Zhang, C. Queral, U. Imke, and V. H. Sanchez-Espinoza, "Multiscale analysis of the boron dilution sequence in the NuScale reactor using TRACE and SUBCHANFLOW," *Nuclear Engineering and Design,* 415 (2023) 112708.

[2] W. J. Choi *et al.*, "Experimental and numerical assessment of helium bubble lift during natural circulation for passive molten salt fast reactor," *Nuclear Engineering and Technology,* 56(3) (2024) 1002-1012.

[3] J. Xie, H. Dong, X. Zhao, and A. Karcanias, "Wind farm power generation control via double-network-based deep reinforcement learning," *IEEE Transactions on Industrial Informatics,* 18(4) (2021) 2321-2330.

[4] S. Aubrun, A. Leroy, and P. Devinant, "A review of wind turbine-oriented active flow control strategies," *Experiments in Fluids,* 58 (2017) 1-21.

[5] P. Garnier, J. Viquerat, J. Rabault, A. Larcher, A. Kuhnle, and E. Hachem, "A review on deep reinforcement learning for fluid mechanics," *Computers & Fluids,* 225 (2021) 104973.

[6] C. Vignon, J. Rabault, and R. Vinuesa, "Recent advances in applying deep reinforcement learning for flow control: Perspectives and future directions," *Physics of fluids,* 35(3) (2023).

[7] M. Kurz, P. Offenhäuser, and A. Beck, "Deep reinforcement learning for turbulence modeling in large eddy simulations," *International journal of heat and fluid flow,* 99 (2023) 109094.

[8] R. S. Sutton and A. G. Barto, *Reinforcement learning: An introduction*. MIT press, 2018.

[9] J. Jeon, J. Lee, and S. J. Kim, "Finite volume method network for the acceleration of unsteady computational fluid dynamics: Non-reacting and reacting flows," *International Journal of Energy Research,* 46(8) (2022) 10770-10795.

[10] J. Jeon, J. Lee, R. Vinuesa, and S. J. Kim, "Residual-based physics-informed transfer learning: A hybrid method for accelerating long-term cfd simulations via deep learning," *International Journal of Heat and Mass Transfer,* 220 (2024) 124900.

[11] L. C. Garaffa, M. Basso, A. A. Konzen, and E. P. de Freitas, "Reinforcement learning for mobile robotics exploration: A survey," *IEEE Transactions on Neural Networks and Learning Systems,* 34(8) (2021) 3796-3810.

[12] B. Hambly, R. Xu, and H. Yang, "Recent advances in reinforcement learning in finance," *Mathematical Finance,* 33(3) (2023) 437-503.

[13] A. Zhang, R. McAllister, R. Calandra, Y. Gal, and S. Levine, "Learning invariant representations for reinforcement learning without reconstruction," *arXiv:2006.10742*, 2020.

[14] D. Wang, R. Walters, and R. Platt, "SO(2)-Equivariant Reinforcement Learning," *arXiv:2203.04439,* 2022.

[15] J. Rabault, M. Kuchta, A. Jensen, U. Réglade, and N. Cerardi, "Artificial neural networks trained through deep reinforcement learning discover control strategies for active flow control,"

*Journal of fluid mechanics,* 865 (2019) 281-302.

[16] J. Rabault and A. Kuhnle, "Accelerating deep reinforcement learning strategies of flow control through a multi-environment approach," *Physics of Fluids,* 31(9) (2019).

[17] V. Belus, J. Rabault, J. Viquerat, Z. Che, E. Hachem, and U. Reglade, "Exploiting locality and translational invariance to design effective deep reinforcement learning control of the 1-dimensional unstable falling liquid film," *AIP Advances,* 9(12) (2019).

[18] H. Tang, J. Rabault, A. Kuhnle, Y. Wang, and T. Wang, "Robust active flow control over a range of Reynolds numbers using an artificial neural network trained through deep reinforcement learning," *Physics of Fluids,* 32(5) (2020).

[19] K. Zeng and M. D. Graham, "Symmetry reduction for deep reinforcement learning active control of chaotic spatiotemporal dynamics," *Physical Review E,* 104(1) (2021) 014210.

[20] M. I. Radaideh *et al.*, "Physics-informed reinforcement learning optimization of nuclear assembly design," *Nuclear Engineering and Design,* 372 (2021) 110966.

[21] C. Li, R. Yu, W. Yu, and T. Wang, "Reinforcement learning-based control with application to the once-through steam generator system," *Nuclear Engineering and Technology,* 55(10) (2023) 3515-3524.

[22] C. Vignon, J. Rabault, J. Vasanth, F. Alcántara-Ávila, M. Mortensen, and R. Vinuesa, "Effective control of two-dimensional Rayleigh–Bénard convection: Invariant multi-agent reinforcement learning is all you need," *Physics of Fluids,* 35(6) (2023).

[23] L. Guastoni, J. Rabault, P. Schlatter, H. Azizpour, and R. Vinuesa, "Deep reinforcement learning for turbulent drag reduction in channel flows," *The European Physical Journal E,* 46(4) (2023) 27.

[24] P. Suárez *et al.*, "Active flow control for three-dimensional cylinders through deep reinforcement learning," *arXiv:2309.02462,* 2023.

[25] S. Peitz, J. Stenner, V. Chidananda, O. Wallscheid, S. L. Brunton, and K. Taira, "Distributed control of partial differential equations using convolutional reinforcement learning," *Physica D: Nonlinear Phenomena,* 461 (2024) 134096.

[26] B. Font, F. Alcántara-Ávila, J. Rabault, R. Vinuesa, and O. Lehmkuhl, "Active flow control of a turbulent separation bubble through deep reinforcement learning," *Journal of Physics: Conference Series*, 2753(1) (2024) 012022.

[27] P. Ladosz, L. Weng, M. Kim, and H. Oh, "Exploration in deep reinforcement learning: A survey," *Information Fusion,* 85 (2022) 1-22.

[28] N. Zolman, U. Fasel, J. N. Kutz, and S. L. Brunton, "SINDy-RL: Interpretable and Efficient Model-Based Reinforcement Learning," *arXiv:2403.09110,* 2024.

[29] T. Cohen and M. Welling, "Group equivariant convolutional networks," *International conference on machine learning*, PMLR (2016) 2990-2999.

[30] M. W. Lafarge, E. J. Bekkers, J. P. Pluim, R. Duits, and M. Veta, "Roto-translation equivariant convolutional networks: Application to histopathology image analysis," *Medical Image Analysis,* 68 (2021) 101849.

[31] A. Vaswani *et al.*, "Attention is all you need," *Advances in neural information processing systems,* 30 (2017).

[32] M. Yu and S. Sun, "Policy-based reinforcement learning for time series anomaly detection," *Engineering Applications of Artificial Intelligence,* 95 (2020) 103919.

[33] Z. Yang *et al.*, "Towards applicable reinforcement learning: Improving the generalization and sample efficiency with policy ensemble," *arXiv:2205.09284,* 2022.

[34] J. Schulman, S. Levine, P. Abbeel, M. Jordan, and P. Moritz, "Trust region policy optimization," in *International conference on machine learning*, PMLR (2015) 1889-1897.

[35] J. Schulman, F. Wolski, P. Dhariwal, A. Radford, and O. Klimov, "Proximal policy optimization algorithms," *arXiv:1707.06347,* 2017.

[36] 2A. Kuhnle, M. Schaarschmidt, and K. Fricke, "Tensorforce: a tensorflow library for applied reinforcement learning," Web page (2017)

[37] M. Schaarschmidt, A. Kuhnle, B. Ellis, K. Fricke, F. Gessert, and E. Yoneki, "Lift: Reinforcement learning in computer systems by learning from demonstrations," *arXiv:1808.07903,* 2018.

[38] J. Kim, P. Moin, and R. Moser, "Turbulence statistics in fully developed channel flow at low Reynolds number," *Journal of fluid mechanics,* 177 (1987) 133-166.

[39] M. Mortensen, "Shenfun's documentation : https://shenfun.readthedocs.io," Web page.

[40] M. Mortensen, "Shenfun: High performance spectral Galerkin computing platform," *Journal of Open Source Software,* 3(31) (2018) 1071.

[41] Z. Li, F. Liu, W. Yang, S. Peng, and J. Zhou, "A survey of convolutional neural networks: analysis, applications, and prospects," *IEEE transactions on neural networks and learning systems,* 33(12) (2021) 6999-7019.

[42] S. Lee and D. You, "Data-driven prediction of unsteady flow over a circular cylinder using deep learning," *Journal of Fluid Mechanics,* 879 (2019) 217-254.

[43] A. Krizhevsky, I. Sutskever, and G. E. Hinton, "Imagenet classification with deep convolutional neural networks," *Advances in neural information processing systems,* 25 (2012).

[44] S. Dieleman, K. W. Willett, and J. Dambre, "Rotation-invariant convolutional neural networks for galaxy morphology prediction," *Monthly notices of the royal astronomical society,* 450(2) (2015) 1441-1459.

[45] T. Cohen, "Equivariant convolutional networks," University of Amsterdam, 2021.

[46] S. F. Stefenon, L. O. Seman, L. S. Aquino, and L. dos Santos Coelho, "Wavelet-Seq2Seq-LSTM with attention for time series forecasting of level of dams in hydroelectric power plants," *Energy,*

274 (2023) 127350.

[47] M. Sanchis-Agudo, Y. Wang, K. Duraisamy, and R. Vinuesa, "Easy attention: A simple self-attention mechanism for transformers," *arXiv:2308.12874,* 2023.